\documentclass[conference]{IEEEtran}
\IEEEoverridecommandlockouts
\usepackage{cite}
\usepackage{amsmath,amssymb,amsfonts}
\usepackage{algorithmic, algorithm}
\usepackage{graphicx}
\usepackage{textcomp}
\usepackage{xcolor}
\usepackage{todonotes}

\usepackage{transparent}
\usepackage{makecell}
\usepackage{multirow}
\usepackage{color, colortbl}
\usepackage[
    colorlinks=true,
    linkcolor=blue,
    filecolor=magenta,      
    urlcolor=blue,
    pdftitle={Overleaf Example},
    pdfpagemode=FullScreen]{hyperref}

\def\BibTeX{{\rm B\kern-.05em{\sc i\kern-.025em b}\kern-.08em
    T\kern-.1667em\lower.7ex\hbox{E}\kern-.125emX}}

\DeclareMathOperator*{\argmax}{arg\,max}

\newcommand{\hlBlue}[2][0.1]{{\transparent{#1}\colorbox{blue}{\transparent{1}#2}}}
\newcommand{\hlRed}[2][0.1]{{\transparent{#1}\colorbox{red}{\transparent{1}#2}}}
\newcommand{\labelNeg}{\textcolor{red}{\textbf{negative}}}
\newcommand{\labelPos}{\textcolor{blue}{\textbf{positive}}}
\newcommand\methodName{XPROAX}

\definecolor{Gray}{gray}{0.9}
\begin{document}

\title{XPROAX-Local explanations for text classification with progressive neighborhood approximation 
}

\author{\IEEEauthorblockN{Yi Cai}
\IEEEauthorblockA{
\textit{L3S Research Center} \\
\textit{Leibniz Universität Hannover}\\
Hannover, Germany \\
cai@l3s.de}
\and
\IEEEauthorblockN{Arthur Zimek}
\IEEEauthorblockA{
\textit{Dept.~of Math.~and Comp.~Science}\\
\textit{University of Southern Denmark}\\
Odense, Denmark \\
zimek@imada.sdu.dk}
\and
\IEEEauthorblockN{Eirini Ntoutsi}
\IEEEauthorblockA{
\textit{Dept.~of Math.~and Comp.~Science} \\
\textit{Freie Universität Berlin}\\
Berlin, Germany \\
eirini.ntoutsi@fu-berlin.de}
}

\maketitle

\begin{abstract}
The importance of the neighborhood for training a local surrogate model to  approximate the local decision boundary of a black box classifier has been already highlighted in the literature. 
Several attempts have been made to construct a better neighborhood for high dimensional data, like texts, by using generative autoencoders. However, existing approaches mainly generate neighbors by selecting purely at random from the latent space and struggle under the curse of dimensionality to learn a good local decision boundary.
To overcome this problem, we propose a progressive approximation of the neighborhood using counterfactual instances as initial landmarks and a careful 2-stage sampling approach to refine counterfactuals and generate factuals in the neighborhood of the input instance to be explained. 
Our work focuses on textual data and our explanations consist of both word-level explanations from the original instance (intrinsic) and the neighborhood (extrinsic) and factual- and counterfactual-instances discovered during the neighborhood generation process that further reveal the effect of altering certain parts in the input text.
Our experiments on real-world datasets demonstrate that our method outperforms the competitors in terms of usefulness and stability (for the qualitative part) and completeness, compactness and correctness (for the quantitative part).
\end{abstract}


\section{Introduction}
With wide applications of machine learning models in real-world scenarios, concerns on transparency of decision making process keep growing \cite{goodman2017european}. 
Explainable artificial intelligence hence becomes a popular aspect that helps human users to understand behaviors of a black box, which hides its internal logic and inner working from users. 
An enormous number of efforts have been put into explaining decision systems working with tabular \cite{plumb2018model, friedman2008predictive} and image data \cite{petsiuk2018rise, selvaraju2017grad, chattopadhay2018grad}. 
Unfortunately, most of these methods cannot be directly applied to text classifiers because of the nature of textual data, thus this field is left underdeveloped.

The main focus of this paper is a local model-agnostic explanation method for text classifiers. 
Explanation methods in this category firstly train a surrogate model with generated neighbors to mimic the local behavior of the black box waiting to be explained, 
and then extract knowledge from such a transparent local predictor as the explanation for an inquired black box decision over an instance. 
Under this circumstance, the quality of generated neighbors dominates the upper-bound of the fidelity that a surrogate model can achieve. 
To generate neighborhood points, perturbation of inputs \cite{strumbelj2010efficient} is a common choice. It modifies the values of numeric features within a small range or switches the values of categorical features into some others. 
Again, this becomes a much more challenging task while dealing with textual data due to the lack of a formal definition of the neighborhood of a text. Some existing works \cite{ribeiro2018anchors, ribeiro2016should} generate neighboring texts by randomly dropping words from the given text. 
XSPELLS \cite{lampridis2020explaining} proposes to generate neighboring texts in a latent space learned through a generative autoencoder. 
This method can generate more realistic (semantically meaningful and grammatically correct) texts than the former choice. 
However, both methods use a random sampling strategy no matter in which space the perturbation is applied, which might not be the optimal option.

In this work, we propose \methodName{}\footnote{The source code is available at \url{https://github.com/caiy0220/XPROAX}} (local e\underline{X}plainer with \underline{PRO}gressive neighborhood \underline{A}ppro\underline{X}imation),
a novel explanation method with progressive neighborhood approximation for text classifiers. 
Main contributions of this work are summarized as follows:
\begin{itemize}
    \item We propose a neighborhood generation method in a latent space for explaining text classifiers with two-staged progressive approximation of neighborhood.
    \item The output of the explanation method is composed of word-level and factual-level explanations for comprehensive understandings.
    \item We design an automatic evaluation method that quantitatively evaluates explanations for text classifiers regarding completeness, compactness, and correctness.
\end{itemize}

The rest of the paper is organized as follows. In Section~\ref{sec:relatedWord}, we discuss state-of-the-art works in the field of black box explanation.
Section~\ref{sec:basics} introduces the problem settings and preliminaries of explaining a text classifier. Afterwards, Section~\ref{sec:mainMethod} details the proposed method \methodName{}. To qualitatively and quantitatively evaluate the proposed method, exhaustive experiments are designed and performed in Section~\ref{sec:evaluation}. Finally, we discuss future research directions and conclude the paper in Section~\ref{sec:conclusion}.

\section{Related work}
\label{sec:relatedWord}
There have been many studies on black box explanation over the past few years \cite{danilevsky2020survey, bodria2021benchmarking}. 
Depending on whether the explanation is input-specific, we can categorize them as: \textit{global} or \textit{local} explanation. 
These explanation methods can be further categorized by the requirements on black box, namely model-agnostic or model-specific. 
As aforementioned, this work falls into the category of local model-agnostic explanation method.

A local model-agnostic method generates explanations for specific instances. 
The instance, which is classified by a black box, is given as the input of the explanation method along with the black box. 
To investigate the decision making process, neighbors around the given instance are generated in the first place by altering feature values \cite{robnik2008explaining, lemaire2008contact}. 
Closer (more similar) neighbors are assigned with higher weights as they are considered to be more representative of the locality and vice versa.
After that, a transparent model (so-called surrogate model) is trained on these generated neighbors with the intention to simulate exact local behaviors of the black box \cite{ribeiro2016should, guidotti2018local}. 
An explanation can then be constructed by analysing the local predictor.

As the first step, the quality of generated neighbors can make a huge difference to the final explanation \cite{laugel2018defining}. Some methods like \cite{ribeiro2016should} and \cite{strumbelj2010efficient} perform a simple random perturbation. 
Basically, random perturbation modifies the values of several randomly selected features within a small range, it iterates the slight fluctuation in the vicinity of given input until sufficiently many neighbors have been generated.
Although this simple approach generates satisfactory neighbors for most data representations, it requires fine-tuning for more complicated forms of data, which is textual data in our case. 
One difficulty of applying perturbation to text is how to modify the value of each feature (word). 
LIME \cite{ribeiro2016should} and SHAP \cite{lundberg2017unified} suggested \textit{word dropping} as a solution -- randomly dropping words from the input text. 
Such an approach is truly intuitive and has been shown to be effective. However, there still remain two problems:
\begin{itemize}
    \item Without considering the internal relationships between words, most of the perturbed instances are incomplete and meaningless. 
    \item The number of possible mutations is limited by the length of the input text. This becomes crucial when the input text is extremely short since the surrogate model will not have enough samples for approximating the black box.
\end{itemize}

Alternatively, XSPELLS \cite{lampridis2020explaining} proposes to generate neighbors of text in a latent space learned through a variational autoencoder \cite{bowman2016generating}. Prior to XSPELLS, ALIME \cite{shankaranarayana2019alime} and ABELE \cite{guidotti2019black} make use of autoencoders to construct a latent space for explaining decisions over tabular and image data respectively.
ALIME still generates neighbors in the original feature space of an input, the autoencoder only serves as a weighing function for computing distances in the latent space.
ABELE deploys an adversarial autoencoder \cite{makhzani2015adversarial} for generating synthetic images. XSPELLS fine-tunes the framework of ABELE so that it can deal with text classifiers. 
It constructs the neighborhood of an input text in a latent space. 
This way, it tackles the limitation of \textit{word dropping} by generating realistic instances with a generative model.
However, the choice of the generative model impacts the final explanations. Ability of reconstruction and a properly organized latent space are highly desired features from the generative model in order to produce neighbors that satisfy the locality constraint. 
Also, a random sampling approach will not be able to efficiently and effectively explore a high-dimensional latent space, so the generation strategy in a latent space is worth further discussing.
Besides, both ABELE and XSPELLS train the local predictor in the latent space.
We argue that building the local predictor in a latent space can be a choice only if there are sufficient investigations on corresponding latent features.

The proposed method extends XSPELLS by addressing the issues mentioned above. It differs from other approaches in the literature because of the progressive approximation of the neighborhood and the diversity-based \mbox{(counter-)}factual selection.


\section{Definitions and preliminaries}
\label{sec:basics}
Let $b(\cdot)$ be a black box model like a Deep Neural Network or a Random Forest. 
We have no access to the internal mechanism of the black box except its decisions for given instances. 
Specifically, for a query text instance $x$, the black box returns the confidence score vector\footnote{For simplicity, we assume a binary classification problem: $y\in \{+, -\}$.}
$b(x)=<p_+,p_->$. The class with the maximum probability is returned to the user, i.e., $\hat{y}=\argmax(b(x))$. 
The goal is to explain the decision of the black box for the given instance $x$, i.e., $b(x)$.
Our method uses the probability vector $b(x)$ rather than only the final decision $\hat{y}$ to comprehensively study the behavior of the black box.

Our approach falls into the category of local model explanation methods which use local surrogate models to approximate the predictions of the underlying black box model. 
These models are built in the ``neighborhood'' of the instance $x$ to be explained. 
Let $N(x)$ denote the neighborhood\footnote{When it is clear from the context, we denote the neighborhood of $x$ by $N$ instead of $N(x)$.} of $x$ and let $m(\cdot)$ be the local surrogate model  trained upon $N(x)$.
For the neighborhood generation, we assume a generative model $G$ that approximates the underlying data distribution (Section~\ref{sec:generativeModel}). 
Contrary to existing approaches that generate neighborhoods by sampling randomly in the latent space \cite{lampridis2020explaining, vlassopoulos2020explaining}, we propose to \emph{progressively} approximate the neighborhood of $x$ using \emph{counterfactual instances-as-landmarks} (Section~\ref{sec:neighborhoodGeneration}). 
The generated instances $N(x)$ are then labeled by the black box model $b(\cdot)$ and comprise the training set for learning the local surrogate model $m(\cdot)$. 
The progressive neighborhood approximation helps us to identify the \emph{turning points} 
for the class decision of $x$ and explain its classification (Section~\ref{sec:explanationsCounterfactuals}) together with information inferred from the neighborhood (Section~\ref{sec:explanationsFeatures}).

\section{\methodName{}: local eXplanations with PROgressive neighborhood ApproXimation}
\label{sec:mainMethod}
A surrogate model approximates the predictions of the underlying black box model locally. In order for the limited capacity model $m(\cdot)$ to be on par with the complex model $b(\cdot)$, the training instances $N(x)$ for building $m(\cdot)$ should be carefully generated locally in the neighborhood of $x$.

To this end, we propose a local model-agnostic explanation method called \methodName{}. The method approximates the neighborhood of the instance to be explained in the latent space (Section~\ref{sec:neighborhoodGeneration}), which is constructed with the locality requirement being taken into account (Section~\ref{sec:generativeModel}). 
Subsequently, we derive the explanations from the constructed neighborhood (Section~\ref{sec:explanationsMain}).

\subsection{Locality-preserving neighborhood generation}
\label{sec:generativeModel}
To build the neighborhood $N$ of the instance $x$ to be explained, we need a mechanism that generates ``meaningful'' neighbors around $x$. A simple idea would be to use corruption~\cite{iosifidis2019sentiment}, which removes or replaces words at random. Nevertheless, this process can result in semantically and/or syntactically invalid texts. 
Generative autoencoders have been successfully used for text generation, however not all models map similar texts to nearby latent vectors \cite{zhao2018adversarially}. 
Therefore, for the generative model we use DAAE \cite{shen2020educating}, an approach that maps similar texts to similar latent representations by augmenting  the adversarial autoencoder with a denoising objective where original texts are reconstructed from their perturbed versions.
DAAE learns neighborhood-preserving text representations while producing high quality texts with properly organized latent space geometry.
Moreover, DAAE retains the ability to reconstruct input texts in contrast to VAE~\cite{bowman2016generating}, a popular text generation method, 
that loses its reconstruction ability due to the KL regularization term \cite{song2019latent, dai2020usual}.
 

The DAAE model, $G$, is trained in an unsupervised manner upon data from the same 
domain, this dataset is denoted by $X_G$;
please note that $X_G$ does not need to include the data which are used for training the black box model $b(\cdot)$.
A randomly sampled subset $X_G^\prime \subseteq X_G$ is retained as a \emph{corpus} for selecting the \emph{initial landmarks}; the corpus is down-sampled for  efficiency but due to random sampling it should still reflect the distribution of the dataset.  
DAAE consists of:
A deterministic encoder $E:\mathcal{X}\rightarrow \mathcal{Z}$ that maps from the text space to the latent space and a probabilistic decoder $D:\mathcal{Z}\rightarrow \mathcal{X}$ that reconstructs word sequences from latent representations. 

To generate the neighbors of a given input $x\in\mathcal{X}$, the generative model $G$ 
will firstly map $x$ to the latent space and will represent it as a latent vector $z=E(x)$. A set of neighboring vectors 
can then be created in the latent space by manipulating the latent vector $z\in\mathcal{Z}$. The manipulation includes pure random perturbation \cite{ribeiro2016should, lundberg2017unified, lampridis2020explaining}, locality-based sampling \cite{laugel2018defining} and neighborhood generation using genetic algorithms \cite{guidotti2018local}.
Clearly, a simple random perturbation cannot lead to a good neighborhood $N(x)$ and consequently, to a good surrogate model $m(\cdot)$.
Reference \cite{laugel2018defining} addresses this problem by sampling instances within a fixed-size hyper-sphere as specified by the closest counterfactual to the query instance $x$. However, within this sphere instances are sampled uniformly. 
Reference \cite{guidotti2018local} uses a genetic algorithm to create a synthetic neighborhood around $x$; however, the employed fitness function only focuses on instances similar to $x$ and moreover, the solution instances might converge to the same or very similar instances over the epochs. 

To overcome the limitations of the above methods and better approximate the local data structure around $x$, we propose a progressive approximation of the neighborhood using landmarks to delimit the neighborhood boundary and by carefully interpolating within this boundary. The initial landmarks are counterfactual instances from the corpus $X_G'$, which are \emph{iteratively} replaced by interpolated counterfactual instances\footnote{Counterfactual instances can be either real instances from the corpus or synthetic instances generated during the latent space exploration.} thus allowing for a refinement of the neighborhood. 
The interpolation is two-staged: i) between counterfactual instances to allow for the exploration of the neighborhood and a larger variety of counterfactuals;
ii) between a counterfactual and the pivot point\footnote{That is, the latent space representation of the input $x$.} $z$ to allow for a more localized approximation of the decision boundary, in contrast to existing approaches that mainly sample uniformly within the neighborhood \cite{ribeiro2016should, lundberg2017unified, lampridis2020explaining}.
Once the neighborhood in the latent space $Z$ is refined, the input instances for building the surrogate model $m(\cdot)$ can be derived using the decoder as: $N=D(Z)$.


\subsection{Neighborhood generation with progressive boundary approximation}
\label{sec:neighborhoodGeneration}
A straightforward approach to local boundary approximation in the latent space would be to use the nearest neighbors with the opposite label of $x$ as being evaluated by the cosine distance, since the distance of vectors in the latent space indicates the dissimilarity of texts \cite{shen2020educating}. Let $\mathit{dist}(z_i, z_j)$ be the cosine distance between two instances $x_i$, $x_j$ with corresponding latent vectors $z_i$, $z_j$. 
While texts are not evenly mapped to the latent space (typically following a Gaussian distribution \cite{bowman2016generating}), a fix range that constraints the neighborhood within a certain hyper-sphere
is not optimal for texts that locate in various manifolds of the latent space.
Therefore, we propose to use some of the preserved corpus $X_G^\prime$ instances as initial landmarks to constrain the locality in the latent space. In particular, we use the \emph{$k$-closest counterfactuals} in the corpus as the \emph{seed landmarks}.
To find these landmarks, the counterfactuals (the instances of $X_G^\prime$ predicted by $b(\cdot)$ as of the opposite label of $x$) are mapped to the latent space via the encoder.
The $k$-closest to $x$ counterfactuals in the latent space comprise the set $C$ of landmark points. 
Intuitively, the landmarks reflect the local distribution of vectors around the input $x$, the usage of landmarks maintains the exploration within the local manifold and results in producing high quality neighbors. 


 Still $C$ depends on the available corpus $X'_G$ and further refinement is possible. We further refine the local neighborhood constrained by the landmarks $C$, using interpolation, a commonly used technique in autoencoder-based text generation \cite{bowman2016generating, song2019latent, zhao2018adversarially} which implements successive and meaningful text manipulations based on two given prototypes  through latent vector operations.
The result of the interpolation between two latent vectors $z_p, z_q$ is given by \eqref{interpolate}:
\begin{equation}
I(z_p, z_q)=\{z_i \mid z_i=z_p+i\cdot \frac{(z_p-z_q)}{s}, 0\leq i\leq s\} \label{interpolate}
\end{equation}
$s$ is the number of steps to be taken during the interpolation. For $s=1$, only one instance is interpolated between $z_p,z_q$.

The neighborhood construction process (Algorithm~\ref{alg:construct}) is an iterative process, in each iteration a better neighborhood approximation is achieved as closer counterfactuals to $x$ are discovered and exploited.
And more neighboring instances are interpolated between $x$ and the discovered counterfactuals (Algorithm~\ref{alg:approximate}).

\begin{algorithm}[tb]
\caption{Neighborhood approximation}
\label{alg:approximate}
\begin{algorithmic}[1]
\renewcommand{\algorithmicrequire}{\textbf{Input:}}
\renewcommand{\algorithmicensure}{\textbf{Output:}}
\REQUIRE $x$: query instance; $C$: set of counterfactual landmarks 
\ENSURE  $N_{new}$: generated neighbors; $C_{new}$: updated $C$
\STATE $C_{new}=\emptyset$
\REPEAT
    \STATE $z_p, z_q = \mathrm{RandomlyDraw}(C, 2)$ \\ \COMMENT{\textit{draw randomly 2 vectors from the landmark set}}
    \STATE $Z^\prime = I(z_p, z_q)$ \\ \COMMENT{\textit{1st interpolation: between counterfactuals}}
    \STATE $Z=\emptyset$
    \FOR[\textit{2nd interpolation: between polarities}]{$z_i\in Z^\prime$}
        \IF{$b(D(z_i)) \neq b(x)$}
            \STATE $Z \leftarrow Z + I(z_i, E(x))$
        \ENDIF
    \ENDFOR
    \STATE $z_c = \mathrm{ClosestCounterfactual}(Z, 1)$
    \STATE $C_{new}.insert(z_c)$
    \STATE $N_{new} \leftarrow N_{new} + D(Z)$
\UNTIL{$k$ times}
\RETURN $N_{new}, C_{new}$ 
\end{algorithmic} 
\end{algorithm}

The neighborhood approximation based on a set of landmarks $C$ is shown in Algorithm~\ref{alg:approximate}.
It takes as input the set of current landmarks $C$ and returns a new set of landmarks $C_{new}$ and a new set of neighboring instances $N_{new}$. 
The \emph{first-stage interpolation} (line 4) takes place between two counterfactuals that are randomly sampled from the landmark set $C$ with replacement (line 3).
For every point created by the first-stage interpolation (line 6), we reconstruct the text from the latent vector and check its label assigned by the black box $b(\cdot)$ (line 7). The generated points $Z'$ can have an opposite label (counterfactuals) or same label (factuals) as $x$.
In the \emph{second-stage interpolation}, we only interpolate between the input point $x$ and the generated points $Z'$ holding the opposite label to $x$ (line 8).
We keep all the neighbors generated through the second stage in a set $Z$ (line 8). 
Subsequently, the new landmark set absorbs the closest counterfactual in $Z$ (line 11, 12), the set of neighbors is expanded by neighboring texts reconstructed from the latent vectors in $Z$ (line 13).
The process is repeated $k$ times (line 2, 14), allowing a better exploration of the constrained local space.

The neighborhood construction process (Algorithm~\ref{alg:construct}) runs iteratively until the number of iterations exceeds the limit or when no closer counterfactuals have been found in successive iterations.  
In each iteration, the method 
updates the set of counterfactual landmarks $C_{new}$ and generates new neighbors $N_{new}$ (line 3) based on the neighborhood  approximation Algorithm~\ref{alg:approximate} (line 2).
Newly generated neighbors are stored as candidates for the final selection (line 4).
After the completion of the construction, we eliminate duplicates contained in $N$ (line 6) and output the $n$ closest factuals and the $n$ closest counterfactuals as the neighborhood $N$ for explaining the decision of $b(\cdot)$ over $x$.

\begin{algorithm}[tb]
\caption{Neighborhood construction}
\label{alg:construct}
\begin{algorithmic}[1]
\renewcommand{\algorithmicrequire}{\textbf{Input:}}
\renewcommand{\algorithmicensure}{\textbf{Output:}}
\REQUIRE $x$: query instance; $C$: initial set of landmarks 
\ENSURE  $N$: neighborhood of $x$
\WHILE{not $\mathrm{terminate}()$}
    \STATE $N_{new}, C_{new} = \mathrm{NeighborhoodApproximation}(C, x)$
    \STATE $C = C_{new}$
    \STATE $N \leftarrow N + N_{new}$
\ENDWHILE
\STATE $N=\mathrm{RemoveDuplicates}(N)$
\STATE $N=\mathrm{Closest}(N, n, n)$ \\ \COMMENT{\textit{output $n$ closest instances for both classes}}
\RETURN $N$ 
\end{algorithmic} 
\end{algorithm}

Through this iterative process a better set of neighboring instances is discovered. 
The neighborhood achieves higher diversity due to the first stage interpolation and highlights the decision boundary between factuals and counterfactuals through the second stage interpolation.

\subsection{Extracting local explanations}
\label{sec:explanationsMain}
The explanations are based on the generated neighborhood $N(x)$. In particular, we propose two explanation components:
i) feature/word-level importance (Section~\ref{sec:explanationsFeatures}) and ii) factual and counterfactual instances identified during the neighborhood approximation process (Section~\ref{sec:explanationsCounterfactuals}).

\subsubsection{Word-level explanation}
\label{sec:explanationsFeatures}
We train the surrogate model $m(\cdot)$ upon the neighborhood $N(x)$ generated in the previous step. 
In particular, we train a linear regression model in the original feature space. 
Unlike \cite{lampridis2020explaining}, we prefer the original feature space over the latent feature space because of the comprehensive understanding on the original features (words), this helps us in deriving knowledge from the surrogate model.
We represent the texts using the bag-of-words model in order to simplify the complexity of the classification task. 
The sequential information vanishes as a trade-off between integrity 
and feasibility. 
The model is trained using the weighted square loss function $\mathcal{L}$ \cite{ribeiro2016should} defined in \eqref{lossfunc}.
The first term in \eqref{lossfunc} indicates the weight of a generated neighbor $x_i$ calculated from its latent space distance to the input $x$ and $\sigma$ is the kernel width controlling the influence of the weighting function. 
The second term is the square loss of the surrogate model $m(\cdot)$ on simulating local behavior of the black box $b(\cdot)$.

\begin{equation}
\label{lossfunc}
\mathcal{L}(N, x)=\sum_{x_i\in N}exp(\frac{-\mathit{dist}(E(x_i), E(x))^2}{\sigma^2})\cdot(b(x_i)-m(x_i))^2
\end{equation}

After training, the weights of features in the regression model can be derived as the importance values of the features/words.
Along with words consisting in the input text, words appearing in the novel neighbors are also rated. 
Therefore, we split this explanation part into two subparts:
i) words in the text $x$ to be explained, we refer to them as \emph{intrinsic features/words}; ii) words that only appeared in the neighborhood $N(x)$, we refer to them as \emph{extrinsic features/words}.

The weights assigned to intrinsic features reveal the most important features in an input that cause the decision. 
Extrinsic features only appearing in the neighborhood are incorporated to facilitate further explanation of the decision \cite{ma2017salient, wang2011image}.
The relationships between the input and the extrinsic words values are indistinct.
For the explanation, however, it is important to ``connect'' extrinsic words to the original instance.
Our idea is to do this by demonstrating how they affect the decision on the input text.
In particular, inspired by deletion on the most important intrinsic words from the input text, we can demonstrate the impact of an extrinsic word on the decision by inserting it into the input text.
Although a random insertion/replacement of word will already affect the decision of $b(\cdot)$, we aim at determining the optimal edition with a given extrinsic word to have a manually created and meaningful \mbox{(counter-)}factual.
This can be achieved by finding a position of insertion/replacement, at which the likelihood of the appearance of an extrinsic word is maximized according to the neighborhood $N$ as stated.
In \eqref{insertion}, we measure the likelihood of a word's appearance as the conditional probability under the local context, which consists of $l$ preceding and $l$ succeeding words. 
The conditional probability of an extrinsic word $w$ can be estimated in the neighborhood.
The variable $i$ denotes a possible position for the edition, the position that maximizes the likelihood will be chosen as the optimal solution. 
To allow the integration of the local contexts with the logarithm, a minute value $\epsilon \ll 1$ is involved as the last part of \eqref{insertion}.
\begin{equation}
\label{insertion}
\argmax_i{\sum_{j=-l}^{l} \log(P(w|w_{i+j})}+\epsilon)
\end{equation}

The example presented in Table~\ref{tbl:editExample} demonstrates a manually created counterfactual through the edition based on an extrinsic word. 
The edition builds the connection between the input ``\emph{would not recommend.}'' and the extrinsic word ``\textit{definitely}'', it reveals that the adverb describing the word \emph{recommend} plays an important role in switching the black box decision.

\begin{table}[tbp]
\caption{An example of the extrinsic word based edition}
\begin{center}
\begin{tabular}{ll} 
\hline
\textbf{Input}: would not recommend . & $b(\cdot)$: \labelNeg\\
\hline
\multicolumn{2}{l}{\textbf{Extrinsic word}: \hlBlue[0.17]{definitely}} \\
\textbf{Edition}: would \underline{definitely} recommend . & $b(\cdot)$: \labelPos\\
\hline
\end{tabular}
\label{tbl:editExample}
\end{center}
\end{table}

\subsubsection{Factuals and Counterfactuals as explanation}
\label{sec:explanationsCounterfactuals}
As a benefit from the generative model, the generated neighbors can also be used as factuals and counterfactuals for further explanation. The neighboring texts illustrate the changes that are relevant or irrelevant to reverse the original prediction result of $b(\cdot)$. 

Instead of focusing on neighbors with minimum latent space distances, which are most likely to generate homogeneous differences to the input text, we take the diversity of the \mbox{(counter-)}factuals into account. 
By having the diversity, it becomes easier to underscore the most important components for the decision even with a limited number of \mbox{(counter-)}factuals, such components tend to remain unchanged in factuals and are most likely to be changed in counterfactuals.
To achieve this, we form a \mbox{(counter-)}factual set by picking one instance at a time until sufficiently many \mbox{(counter-)}factuals are picked. 
Each time, the closer instance that maintains the diversity of the \mbox{(counter-)}factual set will be absorbed. 
Equation \eqref{eqQuality} quantifies the quality of an instance according to our expectation on the \mbox{(counter-)}factual set, we construct the \mbox{(counter-)}factual set $\Xi$ by iteratively picking a neighbor with the maximum $r_i$ value.
\begin{equation}
\begin{split}
r_i=(1-\lambda)\cdot-\mathit{dist}(z_i, z) +\lambda&\cdot\sum^{z_p\neq z_q}_{\Xi\cup\{z_i\}}
\frac{\mathit{dist}(\Delta z_p, \Delta z_q)}{(|\Xi|^2+|\Xi|)/2} \\
&(\text{where}\,\Delta z_i = z_i - z) \label{eqQuality}
\end{split}
\end{equation}
Equation \eqref{eqQuality} consists of two parts: i) the first part measures the distance from a neighbor to the input $x$ in the latent space; 2) the second part measures the normalized diversity \cite{gong2019diversity} of the \mbox{(counter-)}factual set $\Xi$ after absorbing a neighbor $z_i$. 
$\lambda\in[0,1]$ is a user defined parameter to control the importance of the diversity during the construction of the \mbox{(counter-)}factual set.
The first term of \eqref{eqQuality} uses a cosine distance metric and has a negative contribution to the quality, because it is designed to have the closer instances being preferred during the selection.
In the second part, we do not directly measure the diversity among the instances from a constrained neighborhood as it could be subtle.
Instead, we measure the diversity of the difference vectors denoted by $\Delta z_i$, which stands for the difference between the pivot point $z$ and a neighbor $z_i$.
The second part ensures the selected \mbox{(counter-)}factuals manipulate the input in a different way to maintain the diversity.
By greedily picking the neighbor with the highest quality value $r$, the method ensures the diversity of \mbox{(counter-)}factuals. As the explanation set is changing, the quality of a neighbor has to be updated after the absorption of an entry.

\section{Evaluation}
\label{sec:evaluation}
The first goal of our experiments is to qualitatively evaluate the generated explanations in terms of usefulness and stability (Section~\ref{sec:qualitative}).
Secondly, we provide a quantitative evaluation by measuring the confidence drops of the black box predictions after removing important features/words  (Section~\ref{sec:quantitative}). 
Details on the datasets, baselines and parameter selection are provided in Section~\ref{sec:exp_setup}.

\subsection{Experimental setup}
\label{sec:exp_setup}

\textit{Datasets}: We evaluate our approach on two real-world datasets: Yelp reviews \cite{shen2017style} and Amazon reviews polarity \cite{zhang2015character}.
The \emph{Yelp dataset} consists of restaurant reviews, each of which does not exceed 16 words. Reviews are labeled as either positive or negative.
The \emph{Amazon review polarity dataset} consists of  customer reviews about different products with rating ranging from 1 star to 5 stars; \cite{zhang2015character} discarded all neutral reviews (with 3 stars) and re-defined the reviews with $\leq 2$ stars ( $\geq$ 4 stars) as negative (positive, respectively). 
Since we focus on short texts, we only use review titles in our experiments.
We split both datasets into 4 parts: 200K/20K/2K/4K, corresponding to the training set for the generative model and the training/validation/test set for the black box. 
Besides, we used 20K instances randomly drawn from the training set of the generative model as the corpus for initializing the landmark counterfactuals.
The subset for the generative model is notably larger compared to the dataset for the black box model to ensure a good quality generator; this is feasible as DAAE is trained in an unsupervised manner.

\emph{Black box models}: 
As black box models, we used a Random Forest (RF) and a Deep Neural Network (DNN). For the RF, we used the implementation from the \textit{scikit-learn} library, setting the number of weak learners
to 400 and using the TF-IDF vectorizer to transform the raw texts into vectors. 
For the DNN, we used the \textit{Keras} library. We implemented a seven layer DNN; the central layer of the hidden layers is a LSTM \cite{hochreiter1997long}, which captures the sequential information of texts; the rest are fully connected layers with a ReLU activation function.
To train the DNN with texts of various lengths, we transformed raw texts into dense embeddings in the pre-processing stage with a tokenizer and a padder, the padder ensures that all dense embeddings share the same length.

\emph{Neighborhood generation and Surrogate model}:
We trained a DAAE model following~\cite{shen2020educating} as the generative model;
encoder $E$ and decoder $D$ are one-layer LSTMs with hidden dimension 1024 and word embedding dimension 512. 
The encoder finally projects an input $x$ into a latent code $z$ with 128 dimensions for both datasets. 
For the neighborhood approximation, we set the interpolation step $s$ to 10 and the size of the landmark set $k$ to 25. 
These are determined according to our experiments on the effect of parameters\footnote{Due to space limitation the detailed experiments on the effect of parameters are not presented.}. 
The experiments showed that increasing the interpolation step $s$ and the size of landmark set $k$ have a positive impact on the results because it allows to exhaustively explore the space and refine the neighborhood construction.
Our surrogate model $m(\cdot)$ is a linear regression model trained upon the neighborhood $N(x)$.
Details including the performance of the black box and the reconstruction loss $\mathcal{L}_{rec}$ of the generative model are shown in Table~\ref{tbl:dataset}.

\emph{Competitors}: 
We compared our approach against:
\begin{itemize}
    \item \textbf{LIME}~\cite{ribeiro2016should}: the most widely used local explanation method that applies a simple random perturbation to the input $x$ for constructing the neighborhood $N(x)$. 
    \item \textbf{XSPELLS}~\cite{lampridis2020explaining} generates neighbors from the latent space as well and focuses on the textual data.
\end{itemize}

\begin{table*}[tbp]
\caption{Details about datasets, black boxes and generative model}
\begin{center}
\begin{tabular}{|c|c|c|c|c|c|c|c|c|} 
\hline
Dataset & Avg. length & Generative model & \multicolumn{3}{|c|}{Black box} & \multicolumn{2}{|c|}{Accuracy$^{\mathrm{a}}$} & $\mathcal{L}_{rec}$ $^{\mathrm{a}}$\\
\cline{4-6}
\cline{7-8} 
 & of texts & Training Set & Training set & Valid set & Test set & RF & DNN & DAAE \\
\hline
Yelp & 8.83 & 200K & 20K & 2K & 4K & 0.9113 & 0.9548 & 2.79 \\
\hline
Amazon & 8.47 & 200K & 20K & 2K & 4K & 0.7655 & 0.7795 & 3.93 \\ 
\hline
\multicolumn{6}{l}{$^{\mathrm{a}}$The reported values are performance on the test set for black boxes.}
\end{tabular}
\label{tbl:dataset}
\end{center}
\end{table*}



\subsection{Qualitative evaluation}
\label{sec:qualitative}
The evaluation of explanations remains a challenging task with no consent on how it should be conducted.
For the qualitative part, we provide explanations from
different methods on selected instances and assess the explanation quality with human-grounded evaluation \cite{carvalho2019machine, danilevsky2020survey}.
In particular, we analysed the explanations from two different perspectives: 
\begin{itemize}
    \item \textbf{Usefulness}: how the explanations help human understanding on a model decision that has been made.
    \item \textbf{Stability}\cite{carvalho2019machine, alvarez2018robustness}: how similar the explanations for similar instances are.
\end{itemize}

\subsubsection{Usefulness} Table~\ref{tbl:quality} presents the explanations generated by different methods for two different inputs. 
The \textbf{explanation from \methodName{}} consists of four parts: 
i) a saliency map that highlights important \emph{intrinsic words} from the input text $x$ according to the surrogate model $m(\cdot)$,
ii) important \emph{extrinsic words} that only appeared in the neighborhood $N(x)$ (but not in the input) as evaluated by the surrogate model $m(\cdot)$,
iii) top-5 \emph{factuals} from $N(x)$, 
iv) top-5 \emph{counterfactuals} from $N(x)$.
We consider a word to be important if the absolute value of its importance is greater than the threshold 0.1, which holds too for LIME.
The \textbf{explanation from XSPELLS}
also contains four components: i) top-5 factuals, ii) top-5 counterfactuals, iii) the most common words in factuals, and iv) the most common words in counterfactuals. 
The \textbf{explanation from LIME} is represented by a saliency map with the important words in $x$ highlighted.
For \methodName{} and LIME, we illustrate word importance by highlighting important words with colored backgrounds, where the color denotes the class towards which these words contribute. Words highlighted in blue (red) contribute to the positive (negative, respectively) sentiment.
The color intensity indicates how important a word is considered to be by $m(\cdot)$.
Such information is not possible for XSPELLS, since its surrogate model is built in the latent space and therefore no interpretable features can be directly derived from such a model.
Instead, XSPELLS outputs the most frequent words in the \mbox{(counter-)}factuals with their relative frequencies in parenthesis. 
However, these frequencies do not hold direct relation with the surrogate model and the sentiment tendency.


\begin{table*}[tbp]
\caption{Example explanations by different methods for two input instances}
\begin{center}
\begin{tabular}{c | p{15cm} } 
\hline
\rowcolor{Gray}
\textbf{Input 1} & excellent sta for beginning french students . \hfill DNN $b(\cdot)$: \labelPos, Dataset: \textbf{Amazon}\\
\hline
\textbf{\methodName{}} & 
\begin{tabular}{p{7cm} p{7.1cm}}
    \textbf{Saliency}: \hlBlue[0.48]{excellent} sta for \hlBlue[0.09]{beginning} french students . &
    \textbf{Extrinsic words}$^{\mathrm{a}}$: \hlRed[0.35]{worst} \hlRed[0.34]{disappointing} \hlBlue[0.33]{incredible} \\
    \makecell[l]{\textbf{Factuals}: \\
    1) excellent sta for beginning french philosophy. \\
    2) excellent sta for beginning true \_unk\_ students. \\
    3) excellent sta for the beginning war students! \\
    4) excellent sta for beginning french law. \\
    5) an excellent for beginning true \_unk\_ fans.} & 
    \makecell[l]{\textbf{Counterfactuals}: \\
    1) the worst book for the french history.\\ 
    2) not a story for true king \_unk\_ fans.\\
    3) a disappointing for special effects..\\
    4) an disappointing sta for original mystery.\\
    5) sub par for the beginning division fan}
\end{tabular}
\\
\hline
\textbf{XSPELLS} & 
\begin{tabular}{p{7cm} p{7.1cm}}
    \makecell[l]{\textbf{Factuals}: \\
    1) within num days pin\\
    2) another toy book ever\\
    3) old and errors\\
    4) u wan na see \\
    5) at all you may} &
    \makecell[l]{\textbf{Counterfactuals}: \\
    1) atrocious dvd used are \\
    2) nice bra but \\
    3) lousy range only \\
    4) no substance vhs player \\
    5) after num months but why} \\
    \makecell[l]{\textbf{Common words in factuals}:\\
    within (0.067), num (0.067), days (0.067)
    } & 
    \makecell[l]{\textbf{Common words in counterfactuals}:\\
    atrocious (0.083), dvd (0.083), used: (0.083)
    } \\
\end{tabular}
\\
\hline
\textbf{LIME} & \textbf{Saliency}: \hlBlue[0.07]{excellent} sta for \hlBlue[0.05]{beginning} french students .
\\
\hline
\multicolumn{2}{l}{}\\ 
\hline
\rowcolor{Gray}
\textbf{Input 2} & fries are n't worth coming back . \hfill Random Forest $b(\cdot)$: \labelPos, Dataset: \textbf{Yelp} \\
\hline
\textbf{\methodName{}} & 
\begin{tabular}{p{7cm} p{7.1cm}}
    \textbf{Saliency}: fries are n't \hlBlue[0.51]{worth} coming back . & 
    \textbf{Extrinsic words}$^{\mathrm{a}}$: \hlRed[0.66]{not} \hlBlue[0.28]{perfect} \\
    \makecell[l]{\textbf{Factuals}: \\
    1) the fries were n't worth coming. \\
    2) \_unk\_$^{\mathrm{b}}$ are n't worth going back. \\ 
    3) the fries were worth coming back. \\
    4) the fries were worth going back. \\
    5) you do n't be worth coming.} & 
    \makecell[l]{\textbf{Counterfactuals}: \\
    1) \_unk\_ do n't bother in back. \\
    2) \_unk\_ do n't bother going back. \\
    3) \_unk\_ do n't be anybody back. \\
    4) a few fries were definately coming back. \\
    5) \_unk\_ do n't be anybody.}
\end{tabular}
\\
\hline
\textbf{XSPELLS} & 
\begin{tabular}{p{7cm} p{7.1cm}}
    \makecell[l]{\textbf{Factuals}: \\
    1) it seems well they did \\ 
    2) and i feel like on service \\
    3) dave is excellent \\
    4) everything we will get \\ 
    5) all i hung up is nice} &
    \makecell[l]{\textbf{Counterfactuals}: \\
    1) both to die \\
    2) all else s \\
    3) every i may \\
    4) who makes me money last \\ 
    5) all were nt pricey} \\ 
    \makecell[l]{\textbf{Common words in factuals}:\\
    seems (0.091), well (0.091), feel (0.091)
    } & 
    \makecell[l]{\textbf{Common words in counterfactuals}:\\
    die (0.111), else (0.111), every: (0.111)
    }
\end{tabular}
\\
\hline
\textbf{LIME} & \textbf{Saliency}: fries are n't \hlBlue[0.59]{worth} coming back .
\\
\hline
\multicolumn{2}{l}{$^{\mathrm{a}}$Words with high importance that only appeared in the neighbors (not appeared in the input text).}  \\
\multicolumn{2}{l}{$^{\mathrm{b}}$Generic unknown word token for words out of the vocabulary.}  \\
\end{tabular}
\label{tbl:quality}
\end{center}
\end{table*}

The \textbf{first instance} $x$ to be explained is ``\emph{excellent sta for beginning french students.}'' from the Amazon dataset classified as positive by the DNN black box model.
The word ``\textit{excellent}'' is the most important intrinsic word identified by both \methodName{} and LIME, though \methodName{} considers it more important for the positive classification.
The most important extrinsic words by \methodName{} are ``\textit{incredible}'' associated with the positive class and ``\textit{worst, disappointing}'' associated with the negative class.
These extrinsic words are all adjectives which can replace the intrinsic word ``\textit{excellent}'' to generate manually created \mbox{(counter-)}factuals.
Regarding \mbox{(counter-)}factuals, \methodName{} generates meaningful and complete instances with similar sentence structure. The factual ``\emph{an excellent for beginning true \_unk\_ fans.}'' supports the observation from the intrinsic words. Although most of the words in the input text are manipulated, the sentiment remains unchanged with the presence of the important intrinsic words.
On the contrary, XSPELLS suffers from poor (too generic and incomplete) \mbox{(counter-)}factuals like ``\emph{lousy range only}'', which barely make any connection to the input.
So, the explanation by \methodName{} is better and more complete as it covers both intrinsic/extrinsic aspects as well as \mbox{(counter-)}factuals. 
The \textbf{second instance} to be explained is ``\emph{fries are n't worth coming back}'' from the Yelp dataset 
which is classified as positive by the RF black box model, despite the existence of the negation.
Again, XSPELLS does not work properly as most of generated \mbox{(counter-)}factuals are meaningless, e.g. ``\textit{every i may}''. 
Lacking notable relationships with the input, these \mbox{(counter-)}factuals fail in explaining the cause of the decision.
On the other hand, \methodName{} and LIME both perform fairly well. They have an agreement on the primary term that causes the positive decision -- the word ``\textit{worth}''. 
According to this observation, we can already tell that the RF model either has trouble with understanding negation or simply cannot deal with the term ``\textit{n't}''. \methodName{} takes one further step to identify the exact reason.
In extrinsic words, the word ``\textit{not}'' is the highest weighted word in the neighborhood and contributes to a negative sentiment, the observation proves that the RF model is capable to handle the negation under a positive context.
Specifically, the decision can be reversed by replacing the term ``\textit{n't}'' with the extrinsic word ``\textit{not}''.
The other word, ``\textit{perfect}'', has a positive tendency and it is an alternative of the ``positive'' description of the fries. 
Regarding the \mbox{(counter-)}factuals, disappearance of the extrinsic words is a flaw. This is caused by the fact that the neighbors containing extrinsic words are not close enough to the pivot point and therefore are excluded even under the consideration of diversity.
But still, factuals like ``\emph{the fries were worth coming back.}'' reveal the ignorance of the black box on the term ``\textit{n't}'' since there is no obvious influence on classification results regardless of its presence or absence. 

\subsubsection{Stability} An explanation method with higher stability is expected to generate similar explanations for similar inputs. However, there is a lack of a formal definition of the similarity of the input instances, especially for textual data. Intuitively, similar texts should be semantically and syntactically close. For this reason, we manually selected several instances from the test set. To highlight the similarity, we prefer shorter phrases under the same context with exactly the same grammatical structure. 
Table~\ref{tbl:stability} shows the stability of generated explanations by different methods for similar texts. In particular, 
for an input text (the first column), we show in the second column the confidence score of a black box prediction $b(\cdot)$, 
a DNN in this case, in the negative (red color) or positive (blue color) class. 
For \methodName{} and LIME, the importance values of intrinsic words are presented. The number following a word is the importance of the word considered to be for the current decision. 
A positive value means the word results in the current prediction while a negative value leads to the opposite.
XSPELLS does not output word-level importance but only relative frequencies of the most common words (only the top-2 words are presented). 


\begin{table}[tbp]
\caption{Stability of explanations}
\begin{center}
\begin{tabular}{p{16.03mm}c|ccc} 
\multicolumn{5}{c}{Dataset: \textbf{Yelp}, Black box model $b(\cdot)$: \textbf{DNN}}\\
\hline
\textbf{Input Text} & $\pmb{b(\cdot)}$ & \textbf{\methodName{}} & \textbf{LIME} & \textbf{XSPELLS$^{\mathrm{a}}$} \\
\hline
\underline{great} food. & \color{blue}\textbf{1.00}\color{black} & 
\makecell[c]{\underline{great} 0.69 \\ food 0.05} &
\makecell[c]{\underline{great} 0.70 \\ food -0.24} & 
\makecell[c]{food 0.11 \\ italian* 0.08} \\
\hline
\underline{great} sushi. & \color{blue}\textbf{0.99}\color{black} & 
\makecell[c]{\underline{great} 0.62 \\ sushi 0.12} &
\makecell[c]{\underline{great} 0.56 \\ sushi -0.11} & 
\makecell[c]{salsa 0.17\\ guys 0.14}\\
\hline
\underline{great} pizza. & \color{blue}\textbf{1.00}\color{black} & 
\makecell[c]{\underline{great} 0.70 \\ pizza 0.01} &
\makecell[c]{\underline{great} 0.73 \\ pizza -0.27} & 
\makecell[c]{good 0.14\\ food 0.08}\\
\hline
\underline{great} beer. & \color{blue}\textbf{1.00}\color{black} & 
\makecell[c]{\underline{great} 0.64 \\ beer 0.02} &
\makecell[c]{\underline{great} 0.73 \\ beer -0.27} & 
\makecell[c]{good 0.15 \\ story* 0.07}\\
\hline
\multicolumn{2}{l}{}\\ 
\hline
great \underline{food}. & \color{blue}\textbf{1.00}\color{black} & 
\makecell[c]{great 0.69 \\ \underline{food} 0.05} &
\makecell[c]{great 0.70 \\ \underline{food} -0.24} & 
\makecell[c]{\underline{food} 0.11 \\ italian* 0.08} \\
\hline
amazing \underline{food}. & \color{blue}\textbf{0.99}\color{black} & 
\makecell[c]{amazing 0.64 \\ \underline{food} -0.01} &
\makecell[c]{amazing 0.70 \\ \underline{food} -0.24} & 
\makecell[c]{good 0.14 \\ \underline{food} 0.14} \\
\hline
horrible \underline{food}. & \color{red}\textbf{1.00}\color{black} & 
\makecell[c]{horrible 0.45 \\ \underline{food} 0.02} &
\makecell[c]{horrible 0.04 \\ \underline{food} -0.02} & 
\makecell[c]{best* 0.17\\ good* 0.17}\\
\hline
bad \underline{food}. & \color{red}\textbf{1.00}\color{black} & 
\makecell[c]{bad 0.43 \\ \underline{food} 0.02} &
\makecell[c]{bad 0.04 \\ \underline{food} -0.02} & 
\makecell[c]{always 0.29\\ story* 0.17}\\
\hline
\multicolumn{5}{p{8.4cm}}{$^{\mathrm{a}}$XSPELLS outputs the relative frequency of a word, a word without `*' comes from factuals, a word with `*' comes from counterfactuals.} \\
\end{tabular}
\label{tbl:stability}
\end{center}
\end{table}

The DNN model is confident about its decisions since the confidence scores of the predicted class are nearly 1.0.
In the first 4 comments from the table, we focus on the term ``\textit{great}'', it is used in the same context to praise dish/drink. For the explanations from \methodName{} and LIME, the weights of the target term are stable with subtle fluctuation. 
For the last 4 instances, our main focus is the underlined term ``\textit{food}''. In these phrases, the neutral noun is described by different adjectives. \methodName{} generates stable explanations by highlighting the adjectives and assigning small weights to the neutral term. 
In the meantime, LIME fails in producing stable explanations for the decisions that have been made. The importance of the target term varies a lot in the same context. Moreover, although the black box is confident about the negative predictions on the last two examples, LIME considers no words in the last two negative phrases to be important. 
A possible reason that causes such an issue for the perturbation-based explanation method is the size of the training set, which is constructed by the neighboring points. 
For a method that generates neighbors by randomly dropping words, the upper bound of the size of the training set is constrained by the length of the input text. 
Such a shortcoming becomes critical when an input contains only a few words, since the surrogate model cannot be fairly trained with few neighbors and may cause the unstable outputs as presented in the results. 
XSPELLS performs relatively poorly as well. 
Its explanations cannot concentrate on the given input and are less stable.

\textit{Summary}: The above experiments show that \methodName{} generates high quality explanations, which consist of informative intrinsic and extrinsic features/words as well as meaningful \mbox{(counter-)}factuals. 
Our feature/word-level explanations are close to those produced by LIME, but the proposed method generates more stable explanations under the manually selected context. 
In addition, extrinsic words and \mbox{(counter-)}factuals extracted from the carefully constructed neighborhood promote understandings on the black box decisions.
Contrariwise, due to the purely random sampling in the latent space, the \mbox{(counter-)}factuals generated by XSPELLS are not meaningfully related to the input to be explained and consequently, this also holds for the most common words generated from the \mbox{(counter-)}factuals. 
The comparison between \mbox{(counter-)}factuals generated by \methodName{} and XSPELLS intuitively demonstrates the necessity of satisfying locality in the latent space.

\subsection{Quantitative evaluation}
\label{sec:quantitative}
For the quantitative evaluation, we follow the three Cs of interpretability \cite{silva2018towards}, namely \underline{C}ompleteness, \underline{C}ompactness, and \underline{C}orrectness.
Traditionally, the evaluation includes removing words from the input text in the order of the weights assigned by the explanation method and checking the drop of the prediction confidence \cite{lertvittayakumjorn2019human, arras2016explaining}.
Instead of word deletion, we propose to involve the words assigned with negative values in the evaluation.
Same as deleting a positive word (contributing to the predicted class), the insertion of a negative word (contributing to the opposite class) is also expected to cause the drop of confidence in the current prediction.
We named the expanded method as \textit{sentence edition}. The edition process is constrained by setting a minimum threshold $\eta$ for word importance. 
Only words with absolute value above the threshold are considered to be relevant.
Words with positive values are dropped from the sentence while those with negative values are inserted.
For both \methodName{} and XSPELLS, the inserted word can come from the neighbors and the optimal position of insertion is determined by \eqref{insertion}; for LIME, we only perform a random insertion as there is no extra information provided. 

The confidence drop after the explanation-guided edition indicates the \emph{completeness} of an explanation. 
The averaged confidence drop per operation (deletion/insertion) reveals the \emph{compactness}.
For correctness, we consider that weights of features should reflect their contributions to a prediction. 
In other words, deletion/insertion of features with higher weights should have larger impacts on the prediction. 
Therefore, we report the change of compactness $\Delta\eta$ after increasing the threshold $\eta$ as \emph{correctness}. 
A method with higher correctness is expected to generate a more compact explanation after increasing $\eta$ since less relevant words are excluded.
To further reveal the effectiveness of the proposed method and prove that confidence drop cannot be achieved by simply inserting words with strong opposite sentiment, we designed a \textbf{baseline} simulating the behavior of the edition guided by explanations. 
The baseline randomly drops up to 3 words from the input text, meanwhile, it randomly picks a strong sentiment word\footnote{Top-100 words with the strongest sentiment are given by the black box model for both classes.} from the opposite class and inserts the word into the text.

The results of designed experiments are reported in Table~\ref{tbl:effectiveness}.
For confidence drop (completeness) and averaged confidence drop per operation (compactness), the reported values are the mean and the standard deviation of the performance of corresponding method over the test set. 
For the change of compactness ``$\Delta\eta$'' (correctness), the reported value is the change of mean compactness after increasing the threshold.
We empirically set the threshold $\eta$ to 0.1 for measuring completeness and compactness. 
For correctness, we record the change of compactness when we raise $\eta$ from 0.1 to 0.3. 
\methodName{} outperforms the two competitors as well as the baseline in terms of completeness and compactness. In particular, it achieves the highest completeness in all settings and the highest compactness in three settings out of four.
The statistics regarding correctness is in favor of LIME, as the lower compactness by LIME with the threshold $\eta=0.1$ becomes an advantage while evaluating the correctness. 
In other words, it leaves more space for improving the compactness which is used for calculating the correctness.
Nevertheless, \methodName{} still reaches a comparable correctness to LIME.

Admittedly, compared to LIME, \methodName{} benefits from the neighboring extrinsic words.
But the comparison to the designed baseline and XSPELLS demonstrates that \methodName{} achieves better performance based on a better understanding of the black box behavior in the neighborhood as the opposite to simply inserting strong sentiment words from the other class.

\begin{table}[tbp]
\caption{Confidence drop after explanation-guided edition, $\eta=0.1$}
\begin{center}
\begin{tabular}{|c|c|c|c|c|} 
\hline
\makecell{Dataset \\ and \\ Model} & \makecell{Explaining\\ Method} & \makecell{Confidence \\ Drop} & \makecell{Avg. Confidence \\ Drop per op} & \makecell{$\Delta\eta$ \\ (0.3-0.1)}\\
\hline
\multirow{4}{*}{\makecell[c]{Yelp\\ \& \\RF}}
 & baseline & 0.247 ± 0.31 & 0.179 ± 0.24 & /\\
\cline{2-5}
 & LIME & 0.364 ± 0.29 & 0.297 ± 0.26 & \textbf{+0.213}\\
\cline{2-5}
 & XSPELLS & 0.132 ± 0.26 & 0.170 ± 0.27 & +0.032\\
\cline{2-5}
 & \methodName{} & \textbf{0.740 ± 0.22} & \textbf{0.417 ± 0.33} & +0.153\\
\hline
\hline
\multirow{4}{*}{\makecell[c]{Yelp\\ \& \\DNN}}
 & baseline & 0.136 ± 0.32 & 0.094 ± 0.23 & /\\
\cline{2-5}
 & LIME & 0.564 ± 0.46 & \textbf{0.348 ± 0.44} & \textbf{+0.230}\\
\cline{2-5}
 & XSPELLS & 0.084 ± 0.26 & 0.102 ± 0.27 & -0.014\\
\cline{2-5}
 & \methodName{} & \textbf{0.825 ± 0.35} & 0.302 ± 0.43 & +0.206\\
\hline
\hline
\multirow{4}{*}{\makecell[c]{Amazon\\ \& \\RF}}
 & baseline & 0.163 ± 0.19 & 0.118 ± 0.14 & /\\
\cline{2-5}
 & LIME & 0.209 ± 0.18 & 0.201 ± 0.16 & \textbf{+0.185}\\
\cline{2-5}
 & XSPELLS & 0.048 ± 0.13 & 0.058 ± 0.14 & +0.037\\
\cline{2-5}
 & \methodName{} & \textbf{0.506 ± 0.20} & \textbf{0.354 ± 0.21} & +0.126\\
\hline
\hline
\multirow{4}{*}{\makecell[c]{Amazon\\ \& \\DNN}}
 & baseline & 0.287 ± 0.27 & 0.209 ± 0.21 & /\\
\cline{2-5}
 & LIME & 0.424 ± 0.27 & 0.238 ± 0.17 & +0.156\\
\cline{2-5}
 & XSPELLS & 0.095 ± 0.18 & 0.122 ± 0.18 & +0.037\\
\cline{2-5}
 & \methodName{} & \textbf{0.665 ± 0.21} & \textbf{0.298 ± 0.25} & \textbf{+0.164}\\
\hline
\end{tabular}
\label{tbl:effectiveness}
\end{center}
\end{table}


\section{Conclusions}
\label{sec:conclusion}
In this work, we proposed \methodName{}, a local model-agnostic explanation method for text classification that progressively approximates the neighborhood of the instance to be explained and consequently the local decision boundary.
This is achieved by applying a two-staged interpolation in a neighborhood-preserving latent space.
Our explanations consist of word-level importance (including intrinsic and extrinsic words) as well as \mbox{(counter-)}factual instances.

Our experiments, both qualitatively and quantitatively, show the effectiveness and stability of \methodName{} in comparison to state-of-the-art methods. 
Moreover, the experimental results reveal that local explanations on text classifiers must not be limited by the words appearing in a given input (intrinsic words). Neighborhood exploration in a similar context provides a comprehensive view of understanding of the decision that has been made. 
In comparison to XSPELLS, the careful construction of the neighborhood overcomes the weakness of randomly sampling in the latent space. 
By taking advantage of the neighborhood construction, the explanations from our method shows notable relations to the input that XSPELLS cannot guarantee.

As future work, we firstly consider extending the explanation framework for long texts (e.g., documents).
Neighborhood preserving text generation becomes more challenging when the length of text increases.
In this regard, an elegant formatting for input texts and further study on the generative model are essential. 
Second, the choice of the surrogate model remains to be investigated. Although knowledge extraction from a complicated but still transparent surrogate model requires more efforts, accurate simulation on black boxes can be acquired in return.
Finally, \methodName{} is now specialized for textual data, so an interesting future direction is to expand the framework for other forms of data (e.g., image data).

\section*{Acknowledgments}
The first author is supported by the State Ministry of Science and Culture of Lower Saxony, within the PhD program ``Responsible Artificial Intelligence in the Digital Society''. 
We also thank Philip Naumann for the insightful discussions.


\begin{thebibliography}{00}
\bibitem{alvarez2018robustness}
David Alvarez-Melis and Tommi~S Jaakkola.
\newblock On the robustness of interpretability methods.
\newblock {\em arXiv preprint arXiv:1806.08049}, 2018.

\bibitem{arras2016explaining}
Leila Arras, Franziska Horn, Gr{\'e}goire Montavon, Klaus-Robert M{\"u}ller,
  and Wojciech Samek.
\newblock Explaining predictions of non-linear classifiers in nlp.
\newblock In {\em Proceedings of the 1st Workshop on Representation Learning
  for NLP}, pages 1--7, 2016.

\bibitem{bodria2021benchmarking}
Francesco Bodria, Fosca Giannotti, Riccardo Guidotti, Francesca Naretto, Dino
  Pedreschi, and Salvatore Rinzivillo.
\newblock Benchmarking and survey of explanation methods for black box models.
\newblock {\em arXiv preprint arXiv:2102.13076}, 2021.

\bibitem{bowman2016generating}
Samuel Bowman, Luke Vilnis, Oriol Vinyals, Andrew Dai, Rafal Jozefowicz, and
  Samy Bengio.
\newblock Generating sentences from a continuous space.
\newblock In {\em Proceedings of The 20th SIGNLL Conference on Computational
  Natural Language Learning}, pages 10--21, 2016.

\bibitem{carvalho2019machine}
Diogo~V Carvalho, Eduardo~M Pereira, and Jaime~S Cardoso.
\newblock Machine learning interpretability: A survey on methods and metrics.
\newblock {\em Electronics}, 8(8):832, 2019.

\bibitem{chattopadhay2018grad}
Aditya Chattopadhay, Anirban Sarkar, Prantik Howlader, and Vineeth~N
  Balasubramanian.
\newblock Grad-cam++: Generalized gradient-based visual explanations for deep
  convolutional networks.
\newblock In {\em 2018 IEEE Winter Conference on Applications of Computer
  Vision (WACV)}, pages 839--847. IEEE, 2018.

\bibitem{dai2020usual}
Bin Dai, Ziyu Wang, and David Wipf.
\newblock The usual suspects? reassessing blame for vae posterior collapse.
\newblock In {\em International Conference on Machine Learning}, pages
  2313--2322. PMLR, 2020.

\bibitem{danilevsky2020survey}
Marina Danilevsky, Kun Qian, Ranit Aharonov, Yannis Katsis, Ban Kawas, and
  Prithviraj Sen.
\newblock A survey of the state of explainable ai for natural language
  processing.
\newblock In {\em Proceedings of the 1st Conference of the Asia-Pacific Chapter
  of the Association for Computational Linguistics and the 10th International
  Joint Conference on Natural Language Processing}, pages 447--459, 2020.

\bibitem{friedman2008predictive}
Jerome~H Friedman, Bogdan~E Popescu, et~al.
\newblock Predictive learning via rule ensembles.
\newblock {\em Annals of Applied Statistics}, 2(3):916--954, 2008.

\bibitem{gong2019diversity}
Zhiqiang Gong, Ping Zhong, and Weidong Hu.
\newblock Diversity in machine learning.
\newblock {\em IEEE Access}, 7:64323--64350, 2019.

\bibitem{goodman2017european}
Bryce Goodman and Seth Flaxman.
\newblock European union regulations on algorithmic decision-making and a
  “right to explanation”.
\newblock {\em AI magazine}, 38(3):50--57, 2017.

\bibitem{guidotti2019black}
Riccardo Guidotti, Anna Monreale, Stan Matwin, and Dino Pedreschi.
\newblock Black box explanation by learning image exemplars in the latent
  feature space.
\newblock In {\em Joint European Conference on Machine Learning and Knowledge
  Discovery in Databases}, pages 189--205. Springer, 2019.

\bibitem{guidotti2018local}
Riccardo Guidotti, Anna Monreale, Salvatore Ruggieri, Dino Pedreschi, Franco
  Turini, and Fosca Giannotti.
\newblock Local rule-based explanations of black box decision systems.
\newblock {\em arXiv preprint arXiv:1805.10820}, 2018.

\bibitem{hochreiter1997long}
Sepp Hochreiter and J{\"u}rgen Schmidhuber.
\newblock Long short-term memory.
\newblock {\em Neural computation}, 9(8):1735--1780, 1997.

\bibitem{iosifidis2019sentiment}
Vasileios Iosifidis and Eirini Ntoutsi.
\newblock Sentiment analysis on big sparse data streams with limited labels.
\newblock {\em Knowledge and Information Systems}, pages 1--40, 2019.

\bibitem{lampridis2020explaining}
Orestis Lampridis, Riccardo Guidotti, and Salvatore Ruggieri.
\newblock Explaining sentiment classification with synthetic exemplars and
  counter-exemplars.
\newblock In {\em International Conference on Discovery Science}, pages
  357--373. Springer, 2020.

\bibitem{laugel2018defining}
Thibault Laugel, Xavier Renard, Marie-Jeanne Lesot, Christophe Marsala, and
  Marcin Detyniecki.
\newblock Defining locality for surrogates in post-hoc interpretablity.
\newblock In {\em Workshop on Human Interpretability for Machine Learning
  (WHI)-International Conference on Machine Learning (ICML)}, 2018.

\bibitem{lemaire2008contact}
Vincent Lemaire, Raphael F{\'e}raud, and Nicolas Voisine.
\newblock Contact personalization using a score understanding method.
\newblock In {\em 2008 IEEE International Joint Conference on Neural Networks
  (IEEE World Congress on Computational Intelligence)}, pages 649--654. IEEE,
  2008.

\bibitem{lertvittayakumjorn2019human}
Piyawat Lertvittayakumjorn and Francesca Toni.
\newblock Human-grounded evaluations of explanation methods for text
  classification.
\newblock In {\em Proceedings of the 2019 Conference on Empirical Methods in
  Natural Language Processing and the 9th International Joint Conference on
  Natural Language Processing (EMNLP-IJCNLP)}, pages 5198--5208, 2019.

\bibitem{lundberg2017unified}
Scott~M Lundberg and Su-In Lee.
\newblock A unified approach to interpreting model predictions.
\newblock {\em Advances in Neural Information Processing Systems},
  4765--4774, 2017.

\bibitem{ma2017salient}
Ji~Ma, Jingjiao Li, Zhenni Li, and Jiao Jiao.
\newblock Salient region detection by integrating intrinsic and extrinsic cues
  without prior information.
\newblock {\em Journal of Engineering Science \& Technology Review}, 10(3),
  2017.

\bibitem{makhzani2015adversarial}
Alireza Makhzani, Jonathon Shlens, Navdeep Jaitly, Ian Goodfellow, and Brendan
  Frey.
\newblock Adversarial autoencoders.
\newblock {\em arXiv preprint arXiv:1511.05644}, 2015.


\bibitem{petsiuk2018rise}
Vitali Petsiuk, Abir Das, and Kate Saenko.
\newblock Rise: Randomized input sampling for explanation of black-box models.
\newblock {\em arXiv preprint arXiv:1806.07421}, 2018.

\bibitem{plumb2018model}
Gregory Plumb, Denali Molitor, and Ameet Talwalkar.
\newblock Model agnostic supervised local explanations.
\newblock In {\em Proceedings of the 32nd International Conference on Neural
  Information Processing Systems}, pages 2520--2529, 2018.


\bibitem{ribeiro2016should}
Marco~Tulio Ribeiro, Sameer Singh, and Carlos Guestrin.
\newblock "why should i trust you?" explaining the predictions of any classifier.
\newblock In {\em Proceedings of the 22nd ACM SIGKDD international conference
  on knowledge discovery and data mining}, pages 1135--1144, 2016.

\bibitem{ribeiro2018anchors}
Marco~Tulio Ribeiro, Sameer Singh, and Carlos Guestrin.
\newblock Anchors: High-precision model-agnostic explanations.
\newblock In {\em Proceedings of the AAAI Conference on Artificial
  Intelligence}, volume~32, 2018.

\bibitem{robnik2008explaining}
Marko Robnik-{\v{S}}ikonja and Igor Kononenko.
\newblock Explaining classifications for individual instances.
\newblock {\em IEEE Transactions on Knowledge and Data Engineering},
  20(5):589--600, 2008.

\bibitem{selvaraju2017grad}
Ramprasaath~R Selvaraju, Michael Cogswell, Abhishek Das, Ramakrishna Vedantam,
  Devi Parikh, and Dhruv Batra.
\newblock Grad-cam: Visual explanations from deep networks via gradient-based
  localization.
\newblock In {\em Proceedings of the IEEE international conference on computer
  vision}, pages 618--626, 2017.

\bibitem{shankaranarayana2019alime}
Sharath~M Shankaranarayana and Davor Runje.
\newblock Alime: Autoencoder based approach for local interpretability.
\newblock In {\em International conference on intelligent data engineering and
  automated learning}, pages 454--463. Springer, 2019.

\bibitem{shen2017style}
Tianxiao Shen, Tao Lei, Regina Barzilay, and Tommi Jaakkola.
\newblock Style transfer from non-parallel text by cross-alignment.
\newblock In {\em Proceedings of the 31st International Conference on Neural
  Information Processing Systems}, pages 6833--6844, 2017.

\bibitem{shen2020educating}
Tianxiao Shen, Jonas Mueller, Regina Barzilay, and Tommi Jaakkola.
\newblock Educating text autoencoders: Latent representation guidance via
  denoising.
\newblock In {\em International Conference on Machine Learning}, pages
  8719--8729. PMLR, 2020.

\bibitem{silva2018towards}
Wilson Silva, Kelwin Fernandes, Maria~J Cardoso, and Jaime~S Cardoso.
\newblock Towards complementary explanations using deep neural networks.
\newblock In {\em Understanding and Interpreting Machine Learning in Medical
  Image Computing Applications}, pages 133--140. Springer, 2018.

\bibitem{strumbelj2010efficient}
Erik Strumbelj and Igor Kononenko.
\newblock An efficient explanation of individual classifications using game
  theory.
\newblock {\em The Journal of Machine Learning Research}, 11:1--18, 2010.

\bibitem{song2019latent}
Tianbao Song, Jingbo Sun, Bo~Chen, Weiming Peng, and Jihua Song.
\newblock Latent space expanded variational autoencoder for sentence
  generation.
\newblock {\em IEEE Access}, 7:144618--144627, 2019.

\bibitem{vlassopoulos2020explaining}
Georgios Vlassopoulos, Tim van Erven, Henry Brighton, and Vlado Menkovski.
\newblock Explaining predictions by approximating the local decision boundary.
\newblock {\em arXiv preprint arXiv:2006.07985}, 2020.

\bibitem{wang2011image}
Meng Wang, Janusz Konrad, Prakash Ishwar, Kevin Jing, and Henry Rowley.
\newblock Image saliency: From intrinsic to extrinsic context.
\newblock In {\em CVPR 2011}, pages 417--424. IEEE, 2011.


\bibitem{zhang2015character}
Xiang Zhang, Junbo Zhao, and Yann LeCun.
\newblock Character-level convolutional networks for text classification.
\newblock In {\em Proceedings of the 28th International Conference on Neural
  Information Processing Systems-Volume 1}, pages 649--657, 2015.
  
\bibitem{zhao2018adversarially}
Junbo Zhao, Yoon Kim, Kelly Zhang, Alexander Rush, and Yann LeCun.
\newblock Adversarially regularized autoencoders.
\newblock In {\em International conference on machine learning}, pages
  5902--5911. PMLR, 2018.
\end{thebibliography}
\end{document}